\documentclass[wcp]{jmlr}

\usepackage{longtable}

\usepackage{booktabs}

\usepackage{lineno}

\usepackage{colortbl}
\usepackage{mathtools} 
\usepackage{sidecap}
\usepackage{hyperref}       
\usepackage{amsfonts}       
\usepackage{dsfont}
\usepackage{nicefrac}       
\usepackage{microtype}      
\usepackage{xcolor}         
\usepackage{rotating}

\usepackage{amsfonts} 
\usepackage{amsmath}
\usepackage{dsfont}
\usepackage{url}

\newcommand{\R}{\ensuremath{\mathds{R}}}

\newcommand{\parents}{\ensuremath{\mathrm{pa}}}

\usepackage{wrapfig}
\usepackage{xargs} 

\makeatletter
\let\Ginclude@graphics\@org@Ginclude@graphics 
\makeatother

\jmlrvolume{222}
\jmlryear{2023}
\jmlrworkshop{ACML 2023}

\title[Estimation of Counterfactual Interventions under Uncertainties]{Estimation of Counterfactual Interventions under Uncertainties}

  \author{\Name{Juliane Weilbach} \Email{juliane.weilbach@de.bosch.com}\\
  \and
   \Name{Sebastian Gerwinn} \Email{sebastian.gerwinn@de.bosch.com}\\
   \addr Bosch Center for Artificial Intelligence\\
   \Name{Melih Kandemir} \Email{kandemir@imada.sdu.dk}\\
  \addr University of Southern Denmark\\
  \Name{Martin Fraenzle} \Email{martin.fraenzle@informatik.uni-oldenburg.de}\\
 \addr Carl von Ossietzky University of Oldenburg
 }

\editors{Berrin Yan{\i}ko\u{g}lu and Wray Buntine}

\begin{document}
\hypersetup{pageanchor=false}
\editors{Berrin Yan{\i}ko\u{g}lu and Wray Buntine}
\maketitle

\smallskip

\begin{abstract}
Counterfactual analysis is intuitively performed by humans on a daily basis eg. "What should I have done differently to get the loan approved?". Such counterfactual questions also steer the formulation of scientific hypotheses. More formally it provides insights about potential improvements of a system by inferring the effects of hypothetical interventions into a past observation of the system's behaviour which plays a prominent role in a variety of industrial applications.
Due to the hypothetical nature of such analysis, counterfactual distributions are inherently ambiguous. This ambiguity is particularly challenging in continuous settings in which a continuum of explanations exist for the same observation.
In this paper, we address this problem by following a hierarchical Bayesian approach which explicitly models such uncertainty.
In particular, we derive counterfactual distributions for a Bayesian Warped Gaussian Process thereby allowing for non-Gaussian distributions and non-additive noise. 
We illustrate the properties our approach on a synthetic and on a semi-synthetic example and show its performance when used within an algorithmic recourse downstream task.
\end{abstract}

\begin{keywords}
Uncertainty quantification, Counterfactual analysis, Algorithmic recourse 
\end{keywords}

\section{Introduction}\label{sec:intro}
Forming new hypotheses is at the heart of science.
{\em{Counterfactual analysis}} provides essential insights for this task, but also plays an important role in industrial applications. It aims to infer the effects of hypothetical interventions into a past observation of the system's behaviour. Optimizing these interventions for complex decision processes is a key application of counterfactual analysis and is referred to as algorithmic recourse \citep{Karimi21}.
Similarly, such questions are also essential in root-cause analysis \citep{wwwaw2020,budhathoki2022causal} as well as in offline reinforcement learning settings where a control policy needs to be learned only by means of passive observations \citep{buesing2018woulda}. 
However, the graph of the causal dependencies between variables is often not sufficient to perform the necessary optimizations. The analytical or at least computational expressions of functional couplings of the Structural Causal Model (SCM) \citep{pearl2009causal,peters2017elements} are required.
Although these functional couplings can be learnt, they cannot be uniquely identified, due to limited data or intrinsic degrees of freedom of the modelling choices of the functional couplings (see Sec.~\ref{sec:caus-rese-direct} for a particular example). To account for the limited knowledge, Bayesian regression techniques are ideal to keep track of these uncertainties \citep{gal2016dropout,titsias2010bayesian}. When applying causal reasoning, one is often interested in predicting the effect of interventions onto the SCM. This in turn results in queries to the uncertain structural equations with query-points potentially far outside of the training regime. Therefore, using calibrated uncertainties is important for making decisions that are robust under limited knowledge. Here, Gaussian Processes \citep{rasmussen2006gaussian} offer a natural choice to provide calibrated out-of-distribution uncertainties. When using counterfactual distributions to devise recourse actions \citep{Karimi21,ustun2019actionable}, rendering counterfactual explanations more robust against uncertainties, is an active field of research. 
For example uncertainties of functional couplings within a fixed SCM have been analysed by \cite{upadhyay2021towards,dominguez2021adversarial,tsirtsis2021counterfactual,dutta2022robust}. Similarly,  \citep{bui2021counterfactual,dutta2022robust} investigated uncertainties within the classifier. Uncertainty from a disparity between the causal graphs of the data-generating process and the prediction process has been studied by \cite{konig2021causal} and outside of the algorithmic recourse setting by \cite{NEURIPS2021_ca6ab349}, \cite{geffner2022deep} and \cite{NEURIPS2020_0987b8b3}. For a recent survey on this topic, refer to \cite{karimi2021survey}.

Within the setting of counterfactual reasoning, however, there is another inherent uncertainty due to different SCMs featuring disparate parametrization of the stochastic influences yet yielding the same observational as well as interventional distributions (see Figure~\ref{fig:illustration_ambiguous_counterfactual}). Although this non-identifiability of counterfactual distributions can be avoided by imposing additional assumptions onto the underlying structural equation and the exogenous noise distributions \citep{pearl2009causality,shpitser_2007}, these additional assumptions are inherently non-testable and specific modelling assumptions are currently only available for discrete variables within the structural causal model \citep{gumbel_max_oberst19a}. 
Alternatively, identifiability of discrete variable SCMs has been addressed by \cite{DBLP:conf/aaai/ChickeringP96, imbensrubin, richardson11} and \cite{pmlr-v162-zhang22ab} by treating the counterfactual distribution directly as Bayesian variable. However, these approaches are concerned with a discrete setting which allows for direct Bayesian modelling of the resulting counterfactual distributions but do not readily transfer to the continuous setting, which we investigate in this paper.

Instead of imposing non-testable assumptions on the model structure, we propose to follow a hierarchical Bayesian approach which assigns a prior on different parametrizations that leads to potentially different counterfactual distributions and infers the corresponding posterior from observations. By averaging across different parametrizations, we effectively account for all possible counterfactual distributions consistent with the observations.
We equip the established Gaussian Process with random transformations by placing a Normalizing Flow on the likelihood function \citep{transformedGP21}. 
Such a transformation would not only allow for non-Gaussian distributed descendent node variables, but also provide a means to assess possibly different SCMs with the same observational distribution by assigning a probability distribution to different effects of the exogenous noise variable onto the functional coupling.
Using this extended setting, we derive the corresponding counterfactual distribution and show that the resulting distribution over counterfactual estimates can account for non-uniqueness of counterfactual distributions due to ambiguous parametrizations. To evaluate the proposed method in are more realistic setting, we apply it on an established algorithmic recourse benchmark \citep{Karimi21}, thereby assessing the impact of the counterfactual distribution on the downstream task as making accurate decisions in such settings based on quantitative results requires handling uncertainties effectively. 
Our contributions can be summarized as follows: We present a method which allows us to
(i) capture {\it{uncertainty about the parametrization of an SCM}} additionally to the uncertainty in the functional couplings and exogenous noise uncertainty about continuous variables; (ii) derive a counterfactual distribution in this extended setting and (iii) investigate the impact of modelling additional uncertainties on an important downstream task of algorithmic recourse. 

\section{Background and notation}\label{sec:notation}
In this section, we recap relevant concepts of the causal inference literature, including structural causal models, counterfactuals and algorithmic recourse.
\label{sec:scm}
\begin{definition}[Structural causal model\citep{pearl2009causality}\label{def:scm}]
A structural causal model $\mathcal{M}=({\bf{S}},P_U,\mathcal{G})$ is defined via structural equations $S_r, r=1\dots,d$:
\[S_r: X_r = f_r(X_{\rm{\small{pa}}(r)}, U_r)\] 
describing the functional relationship between observational variables $X_r$, for $r=1,\dots,d$. The dependence structure of these variables is defined via an acyclic graph $\mathcal{G}$ determining the parents $\rm{pa}(r)$ of a node $r$. Within the structural equations, exogenous noise variables $U_r$ influence the stochastic assignment of the observational variables and in turn are distributed independently according to $U_r\sim P_{U_r}$.
\end{definition}
Throughout the paper, we assume that there is no latent confounder influencing multiple observational variables, reflected by the independence assumption of $P_{U}=\prod_r P_{U_r}$ of the exogenous variables $U_r$. For a given SCM, observations $X=(X_1,..,X_d)$ can be generated by sampling $U_r\sim P_{U_r}$ and subsequently applying the functional couplings. 
To incorporate imperfect knowledge into the notation of an SCM, we extend Definition~\ref{def:scm} to allow functional couplings to be subject to further uncertainty:
\begin{definition}[Uncertain SCM]
An uncertain structural causal model $\mathcal{M}=({\bf{S}},P_{F},P_{U},\mathcal{G})$ additionally contains a distribution $P_F$ which allows for specifying independent distributions over functional couplings within an SCM $\mathcal{M}$:
\[S_r: X_r = f_r(X_{\rm{\small{pa}}(r)}, U_r), \quad U_r \sim P_{U_r}, f_r\sim P_{F_r}\] 
\end{definition}
Note that, within the above definition, we introduced an additional distribution over functional couplings $f_r$ without increasing the expressiveness of the SCM, however, it allows us to separately interpret different random effects: exogenous noise and imperfect knowledge of functional mappings. To estimate interventional distributions, we would average across both random influences, whereas for counterfactual analysis, we fix the exogenous noise influence and only average across our imperfect knowledge of the functional mappings. With a slight abuse of notation, we do not distinguish between uncertain and deterministic SCMs as deterministic SCMs are a special case of uncertain SCMs by defining a point mass distribution $P_F$ on the deterministic functional couplings. Within an SCM, we denote interventions using the {\em{do}}-operator. That is, when intervening on a set of variables $X_\mathcal{I}=(X_{\mathcal{I}_1},\dots X_{\mathcal{I}_a})$ to set values ${\bf{\theta}}$ explicitly for these variables, we substitute the corresponding structural equations by $S_{\mathcal{I}_i}: X_{\mathcal{I}_i}=\theta_i$ and denote the corresponding derived SCM with $\mathcal{M}[{do(X_\mathcal{I}={\bf{\theta}})}]$. With $P_{\mathcal{M}}$ we denote the data-generating distribution from which observations $X$ can be generated by propagating samples of $X_{\rm{\small{pa}}(r)}$ to $X_r$ via sampling $U_r$ and applying the functional mapping.

\paragraph{Counterfactuals}\label{sec:counterfactual}
\newcommand{\factum}{\ensuremath{{X}^F}}
Counterfactual analysis estimates hypothetical alternative outcomes that would arise if an individual had made a different decision. It is therefore directly linked to a particular observation $\factum$ generated from the underlying SCM (see Definition~\ref{def:scm}).
To perform this kind of analysis, in a first {\em{abduction}} step \citep{pearl2009causality}, a noise posterior distribution $P_{U|\factum}$ is calculated. This noise posterior distribution restricts the exogenous noise influences to the ones which are consistent with the given factum $\factum$ within the functional couplings of a given SCM $\mathcal{M}$.
Consequently, for a given SCM $\mathcal{M}$ and factum $\factum$, we denote the adapted counterfactual SCM $\mathcal{M}_{|\factum} = ({\bf{S}}, P_{U|\factum}, \mathcal{G})$. 
Calculating the noise posterior depends on both the functional coupling $f_r$ and the noise distribution $P_U$, which is particularly challenging when functional couplings are also considered to be probabilistic, i.e., $f_r \sim P_{f_r}$. Within this paper we rely on available results for calculating noise posterior distribution for the case in which functional couplings and noise distributions are modelled with Gaussian distributions, which we state in the following:
\begin{proposition}[Noise posterior of a Gaussian Process \citep{Karimi21}\label{prop:NP-GP}] \newcommand{\ytrain}{{\bf{X}}_r}
\newcommand{\xtrain}{{\bf{X}}_{\rm{\small{pa(r)}}}}
\newcommand{\yfactum}{{x_r^F}}
\newcommand{\xfactum}{{x_{\rm{\small{pa}}(r)}^F}}
Let a node $r$ of an uncertain SCM in which the functional couplings are distributed according to a Gaussian Process with kernel $k_r$ and additive noise $U_r$ be given by:
\[X_r =  f_r(X_{\rm{\small{pa}}(r)}) +  U_r;f_r \sim \mathcal{GP}(0,k_r); U_r\sim \mathcal{N}(0,\sigma_r^2)\]

For an observed factum $\factum$ with $x_r^F, x_{\rm{\small{pa(r)}}}^F$ containing descendent and parent observations according to the graph $\mathcal{G}$ and training data ${\bf{X}}_r=\{x_r^i\}_{i}, {\bf{X}}_{\rm{\small{pa(r)}}}=\{X_{\rm{\small{pa(r)}}}^i\}_{i}$, the noise posterior $P_{U|\factum}$ is given by:

\begin{small}
\begin{align*}
\begin{split}
&P(U_r|\factum) = \mathcal{N}(\mu_r, \Sigma_r), \,\,  {\text{with}}\quad \mu_r =\sigma_r^2\left(\left(K_r + \sigma^2_r \mathds{1}\right)^{-1}\left(\ytrain, \yfactum \right)\right)_{i_F},\\
& \Sigma_r =\sigma_r^2 \left(\mathds{1} - \sigma_r^2 \left( K_r +\sigma_r^2 \mathds{1} \right)^{-1}\right)_{i_F,i_F},\,
K_r = k_r\left(\left(\xtrain, \xfactum \right), \left(\xtrain, \xfactum \right)\right)
\end{split}
\end{align*}
\end{small}
where $i_F$ indicates the index of the factum, i.e. the last entry, as training data and factum are concatenated.
 \end{proposition}

\paragraph{Algorithmic recourse}\label{sec:recourse}
The algorithmic recourse setting aims at finding a counterfactual explanation\citep{wachter2017counterfactual}, which would have led to a more desirable outcome for a particular individual represented by observations $\factum$. This can be translated into an optimization problem in which the outcome is characterized by a given classifier $h:\mathcal{X}\rightarrow [0,1]$ 
from which the outcome of an observation, e.g., getting a loan approved, can be predicted by thresholding $h(x)\geq 0.5$ or alternatively sampled according to the probability $h(x)$. In turn, the recourse problem can be formulated as a constrained optimization problem which minimizes the costs for performing an intervention under the constrain that it would have led to an alternative (more desirable) outcome. The costs are typically associated with the distance of the action of setting a particular value to the factum for which one would like to obtain a different outcome, as performing such action would require to change the individual or its properties. In \cite{Karimi21}, it is extended to also account for the uncertainty within the functional couplings resulting to the following algorithmic recourse formulation:

\begin{align}
    \label{eq:algorithmic_recourse_uncertainty}
    \min_{a=do({\bf{X}_\mathcal{I}})}  {\text{cost}}(a,\factum)\, &s.t. P_{\mathcal{M}[a]_{|\factum}}(h(X))\geq 1-\delta
\end{align}

Eq.~\ref{eq:algorithmic_recourse_uncertainty} minimizes the cost of an action $a$ (performing interventions on an intervention set $\mathcal{I}$) for an individual  $X^{F}$ (one observation, "negatively" classified) such that the found counterfactual sample $X$ reaches the "positive" side after being applied to a classifier $h$ under the counterfactual distribution ($P_{\mathcal{M}[a]|X^F}$). Herein $\delta$ specifies the residual risk that one is willing to accept for not achieving the desired outcome. The constraint therefore measures the minimal probability which can be stated as a threshold on the expectation of the classifier. Note that the constraint in the above optimization problem is specified in terms of the counterfactual distribution. In this paper, however, instead of requiring a high success rate under a single counterfactual SCM, we additionally average across possible SCMs, i.e. replacing the constraint in Eq.~\ref{eq:algorithmic_recourse_uncertainty} by $P_r\left(P_{\mathcal{M}_\phi[a]_{|\factum}}(h(X))\right)\geq 1-\delta$. Here, $P_r$ represents the distribution over possible $\phi$-parametrized SCMs $\mathcal{M}_\phi$ that are all consistent with the observations.
By introducing additional uncertainty which only affects the counterfactual distribution, we expect a more uncertain classification outcome under the counterfactual distribution and hence also expect more robust recourse actions.

\section{Method}
\label{sec:caus-rese-direct}

\begin{wrapfigure}{r}{0.7\textwidth} 
    \centering
    \includegraphics[width=0.7\textwidth]{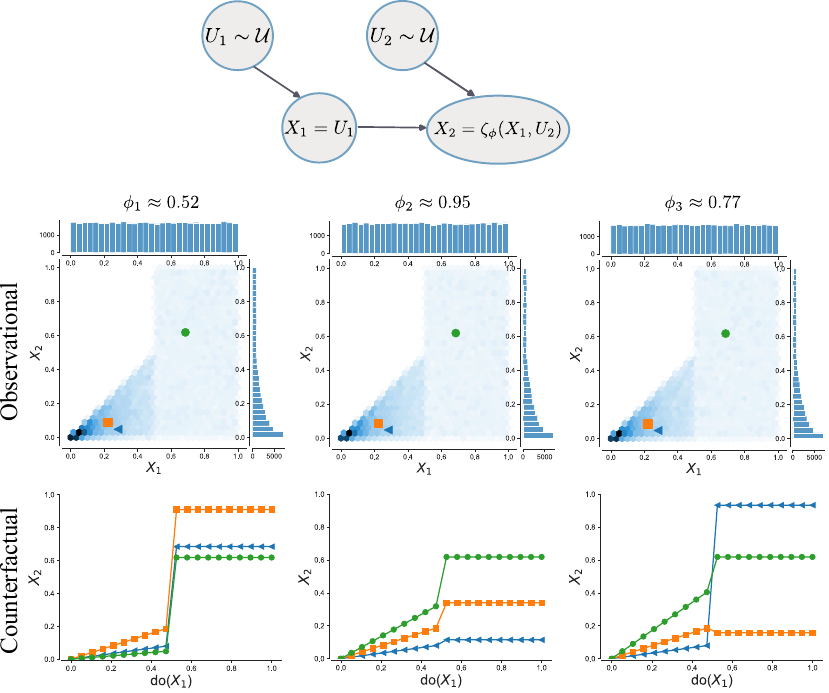}
      \caption{Illustration of the SCM in Eq.\ref{eq:illustrative_example_scm}}
    \label{fig:illustration_ambiguous_counterfactual}
\end{wrapfigure}

The interventional or observational distribution of an SCM are determined by the conditional distributions $p(X_r|X_{\rm{\small{pa}}(r)})$. These distributions, however, can be realized with different combinations of functional coupling and exogenous noise influences. The chosen representation determines the counterfactual distribution in which the exogenous noise influence is kept fixed.
To illustrate this effect of different parametrizations of the noise influence and functional coupling, consider the following adapted example from \cite{peters2017elements}. We construct a family of SCMs $\mathcal{M}_\phi$ with $\phi\in [0,1)$, over two observational variables $X_1, X_2$. All members of the family give rise to the same observational and interventional distributions, but each leads to different counterfactual distributions. Specifically, these SCMs are constructed using the following relationship between $X_1, X_2$ and the corresponding noise influences $U_1, U_2$:

\begin{align}
    X_1 & = U_1, \quad U_1 \sim \mathcal{U}[0,1] ; \quad     X_2  = \mathds{1}_{X_1<0.5}U_2X_1 + \mathds{1}_{X_1\geq 0.5}\zeta_\phi(U_2) \quad U_2 \sim \mathcal{U}[0,1]\nonumber\\
    \zeta_\phi(u) & = \mathds{1}_{u+\phi \geq 1}(u+\phi-1) + \mathds{1}_{u+\phi<1}(u+\phi) \label{eq:illustrative_example_scm}
\end{align}
Here, $\zeta_\phi$ modifies a uniform distribution $\mathcal{U}[0,1]$ by shifting its support by $\phi$ and re-mapping it to $[0,1]$ by cutting off all values larger than 1 and mapping them to $[0,\phi]$. Consequently, the resulting random variable shares the same cumulative distribution function as $\mathcal{U}[0,1]$. However, solving for a particular realization $u$ for a given factual observation $(x_1, x_2)$ results in the following dependence on $\phi$:

\begin{align*}
& u_1  = x_1; & u_2  = \left\{\begin{array}{lr} \frac{x_2}{x_1} &  x_1 <0.5 \\  \frac{x_2}{x_1} -\phi + \mathds{1}_{{x_2} < \phi x_1} & x_1 \geq 0.5 \end{array}\right.
\end{align*}
That is, depending on the value of $x_1$ we either observe a reparametrized version of $u_2$ or $u_2$ directly. In particular, if $x_1<0.5$ is observed, the noise posterior is independent of the parametrization, yet different parametrization will lead to different interventional predictions when intervention are applied in the $x_1>0.5$ regime. Due to this dependence, all these SCMs have different counterfactual distributions, as illustrated in Fig.~\ref{fig:illustration_ambiguous_counterfactual}.
 Here, the  graphical causal model on the left contains a free parameter $\phi$ that characterizes the way the exogenous noise influence affects the SCM. The parametrization is chosen such that each evaluation of such a representational parameter $\phi_1,\phi_2,\phi_3$ leads to the same observational distribution of $X_1,X_2$ (top row in Fig..~\ref{fig:illustration_ambiguous_counterfactual}) when marginalizing out $U_1,U_2$. 
      The conditional of this observational distribution $p(X_2|X_1)$ coincides with the interventional distribution $p(x_2|do(X_1)=x_1)$ due to the simple generating SCM in which $X_1$ corresponds to the root-node. For the three different observations indicated by the markers in the top row, we constructed the counterfactual distributions (three lines, bottom row) for three different representational parameters $\phi_1,\phi_2,\phi_3$. Although the observational distributions are identical, the bottom row shows  different counterfactual distributions corresponding to the 3  SCMs.
As the observational distributions are identical across all parametrizations, the exact SCM cannot be recovered even if infinite amount of data were available. In practice, this is further complicated as only a limited amount of data can be obtained, from which the functional couplings and interaction with the exogenous noise distributions would have to be estimated.

\newcommand{\Y}{{{\bf{\ensuremath{X}}}}_r}
\newcommand{\X}{{{\bf{X}}}_{\parents(r)}}
To this end, consider $i=1,\dots,N$ observations $(x_r)_{i}, \Y=\{(x_r)_i\}_{i=1\dots,N}$ from a node $r$ as well as the corresponding observations $\X = \{(x_{\parents(r)})_i\}_{i=1,\dots,N},(x_{\parents(r)})_i\in \R^{\text{dim}(\parents(r))}$ from the parent nodes $\parents(r)$. To model their relationship, we use the following generative model:

\newcommand{\y}{\ensuremath{X_r}}
\newcommand{\x}{X_{\parents(r)}}
\begin{definition}[Bayesian Warped GP (BW-GP)]
Given kernel parameter $\theta$ and a distribution over parametrizations $p_\phi$, we refer to the following as a Bayesian Warped GP:
\begin{align}
\begin{split}
\label{eq:generative_model}
    &X_r = g_{\phi}^{-1}(f(X_{pa(r)}) + U_r, X_{pa(r)}), \quad f  \sim \mathcal{GP}(\mu_{\mathcal{GP}}, k_{\theta^r}),\quad    U_r \sim \mathcal{N}(0,\sigma_r), \quad \phi  \sim p_\phi
\end{split}
\end{align}
Here, $g_\phi$ is a parametrized mapping, in this paper modeled by a Normalizing flow, which is bijective w.r.t. $\y$ for all $\x$. This renders the model similar to the post-nonlinear causal model \citep{postnonlin12}. The possible parametrizations within the model are represented by the Bayesian belief $p_\phi$. 
\end{definition}
 
 Note that a BW-GP is equivalent to a transformed Gaussian Process with $\mathbb{G}=\mathbb{I}, \mathbb{T} = g_\phi$ within the notation of \cite{transformedGP21}. By inverting the bijective mapping $g_\phi$ w.r.t. its first argument we transform the likelihood (not the prior) of a Gaussian Process. As $g_\phi$ is non-linear  $X_r = g_\phi^{-1}(f(X_{pa(r)}) + U_r, X_{pa(r)})$ is {\it{non-Gaussian}} with {\it{non-additive noise}} \citep{transformedGP21}. By allowing for a non-linear warping using a Normalizing flow, this Gaussian distribution can be mapped to any other distribution of the same dimension arbitrarily well (under some mild regularity assumption see \citep{Koehler20}), provided that the neural network is sufficiently flexible. 
To learn such a model, we employ mean field variational inference. More precisely, using $q_\phi = \mathcal{N}(m,\text{diag}(s))$ as a variational approximation to the true posterior $p_\phi(\cdot |\X, \Y,\theta )$, we optimize the following stochastic approximation (using $S$ samples) to the evidence lower bound (ELBO) \citep{transformedGP21}:

\begin{small}
\begin{align}
 \mathcal{L}(m,s,\theta) &= \mathbb{E}_{q_\phi}\left[\log\left(p(\Y|\X,\phi,\theta)\right)\right] - \text{KL}\left[q_\phi|| p_\phi\right] \nonumber \\
         & \label{eq:approx_elbo}
    \approx \frac{1}{S} \sum_{\phi_i \sim q_\phi} \log\left(p(\Y|\X,\phi_i,\theta)\right)- \text{KL}\left[q_\phi || p_\phi\right]
\end{align}
\end{small}
Here, the marginal likelihood for a fixed transformation $g_\phi$ is given by (see also \citep{snelson_warped_gp}):
\begin{align*}
    &\log\left(p(\Y|\X,\phi,\theta)\right)  = \frac{1}{2} \log \left|\bf{K_\theta} \right| +\frac{1}{2}\mathbf{z}^\top \bf{K_\theta}^{-1} \mathbf{z} - \sum_i \log\left|\frac{\partial g_{\phi}}{\partial x^r}\left({x_i^r}, x_i^{\parents(r)}\right)\right| + \frac{N}{2}\log (2 \pi),\\ 
& {\text{with}}\quad {\bf{K}}_{\theta^r} = \left(k_\theta\left(\X,\X\right) +\sigma \mathds{1}\right);\quad 
\mathbf{z} = \left(g_{\phi}\left(\Y,\X\right) - \mu_{\mathcal{GP}}\left(\X\right)\right)
\end{align*}

The ELBO in Eq.~\ref{eq:approx_elbo} is a lower bound on the observational data distribution as a function of the parameters $m$, $s$, $\theta$, where $m$ and $s$ are the mean and variance of the variational approximation $q$, $\phi$ whereas $\theta$ summarizes parameters from the Gaussian process and therefore enter the first likelihood term only.
Once we have obtained an approximate posterior distribution $q_\phi$ and kernel parameters $\theta$ by optimizing the ELBO Eq.~\ref{eq:approx_elbo}, we can also perform predictions using the generative model Eq.~\ref{eq:generative_model}. Specifically, as the generative model is a Gaussian Process for any fixed transformation within the transformed space, we first sample parameters $\phi\sim q_\phi$. Using this fixed transformation, we can sample a function and noise values on any given test input and transform the sampled observation back into the original space \citep{snelson_warped_gp}.

The resulting process is a hierarchical Bayesian model in which the distribution $q_\phi$ determines the different noise-parametrizations and conditioned on this transformation, the residual uncertainty associated with limited amount of data is captured by a Gaussian Process. In \cite{Karimi21} Gaussian Processes have also been used to model an SCM under imperfect knowledge. This allows for calculating counterfactual distributions and hence enables to analyse the potential outcome of a different decisions even when the functional couplings between the causal variables are not fully known. However, Gaussian Processes fail to model non-Gaussian exogenous noise distributions for transitions between two causally linked variables $X\rightarrow Y$. 

In contrast, Normalizing Flows \citep{papamakarios2021normalizing} offer an alternative which can model complex densities while maintaining analytical tractability for density evaluation and sampling. Combining Gaussian Process with normalizing flow has already been pursued in \cite{transformedGP21}. However, they have not previously been used for the purpose of calculating counterfactual distributions. Exploiting the Gaussian Process property for a fixed transformation in the hierarchical Bayesian model, we can use and extend the result Prop.~\ref{prop:NP-GP} on calculating counterfactual SCMs for GPs to derive a sampling procedure for the counterfactual distribution of a Bayesian warped GP.

\newcommand{\dd}{\mathrm{d}}
\begin{proposition}[Noise posterior distribution of a BW-GP\label{prop:NP-BW-GP}]
Let $\mathcal{M}$ be an uncertain SCM in which the functional couplings are distributed according to a BW-GP. 
\newcommand{\ytrain}{{\bf{X}}_r}
\newcommand{\xtrain}{{\bf{X}}_{\rm{\small{pa(r)}}}}
\newcommand{\yfactum}{{x_r}^F}
\newcommand{\xfactum}{{x_{\rm{\small{pa}}(r)}}^F}

For an observed factum $\factum$ with $x_r^F, x_{\rm{\small{pa(r)}}}^F$ containing descendent and parent observations according to the graph $\mathcal{G}$, training data ${\bf{X}}_r, {\bf{X}}_{\rm{\small{pa(r)}}}$, the noise posterior is given by:

\begin{align} 
\begin{split}\label{eq:noise_posterior}
&P(U_r|\factum) = \int_\phi \mathcal{N}(\mu_r(\phi), s_r(\phi)) q_\phi(\phi) \dd \phi, \,\,
{\text{with}}\\ 
& \mu_r(\phi)  =\sigma_r^2\left({\bf{K}}_r \left(g_\phi({\bf{Y}},{\bf{X}}) -\mu_{\mathcal{GP}}({\bf{X}} \right) \right)_{N+1}; \quad s_r(\phi) =  \sigma_r^2 \left(\mathds{1} - \sigma_r^2 {\bf{K}}_r\right)_{N+1,N+1}\\
& {\bf{K}}_r = \left(k_{\theta^r}({\bf{X}}, {\bf{X}}) + \sigma^2_r \mathds{1}\right)^{-1};\quad
 {\bf{X}}= (\xtrain, \xfactum ); {\bf{Y}}= (\ytrain,\yfactum)
\end{split}
\end{align}
where $N+1$ is the last entry, i.e., the index of the factum when concatenated with the training data $\xtrain,\ytrain$.
\end{proposition}

\paragraph{Proof}
\newcommand{\ytrain}{{\bf{X}}_r}
\newcommand{\xtrain}{{\bf{X}}_{\rm{\small{pa(r)}}}}
\newcommand{\yfactum}{{x_r}^F}
\newcommand{\xfactum}{{x_{\rm{\small{pa}}}(r)}^F}
The statement follows from the fact that for a given transformation, which is specified by $\phi$, $g_\phi((\ytrain,\yfactum),(\xtrain, \xfactum))-\mu_{GP}((\xtrain, \xfactum))$ is distributed according to a zero-mean Gaussian with covariance given by $k_{\theta^r}((\xtrain, \xfactum ), (\xtrain, \xfactum))$. The rest follows by applying Prop.~\ref{prop:NP-GP}. \newline

 Equation\eqref{eq:noise_posterior} also directly gives us a way to approximate the noise posterior by first sampling $\phi$ from the variational approximation $q_\phi$ and subsequently sampling a latent function and corresponding observational noise. To sample from the counterfactual distribution, similarly to \cite{Karimi21}, we average across latent functions, but also across different parametrizations as modelled by $p(\phi)$. Specifically, by exploiting $p(f_r(x^*), U_r|\phi, x^*,{\factum}) = p(f_r(x^*)|\phi,x^*,\factum) p(U_r|\phi,\factum)$, we can first sample from the predictive distribution of the BW-GP and add a sample from the noise distribution according to Eq.\eqref{eq:noise_posterior} in order to get a sample from the counterfactual distribution in which we intervened on the parent node of $r$ and estimate its effect for the observed factum $\factum$. Note, that the noise posterior depends on the transformation $\phi$ only via the transformed values for the descendant nodes. Consequently, the variance and especially the inverse of the kernel matrix can be computed beforehand and independently for all samples of the counterfactual distribution. Calculating the counterfactual distribution for the BW-GP only requires averaging across additional samples for the parameters of the normalizing flow for which also the training data has to be transformed. Consequently, although most causal reasoning methods, including algorithmic recourse do not scale well due to the large number of possible intervention sets, the present method only adds linear computation effort compared to the GP-SCM due to the additional samples of different parametrizations.

\section{Experiments}\label{sec:experiments}
In the following, we evaluate our Bayesian Warped GP model (Eq~.\ref{eq:generative_model}) on the illustrative example (Eq. \ref{eq:illustrative_example_scm}) as well as on a algorithmic recourse benchmark. 
In these experiments, we represent the bijective mapping $g_\phi$  by a neural spline flow with element-wise (referred to as bins) rational conditional spline functions \citep{durkan2019neural,pmlr-v108-dolatabadi20a} and use an independent normal prior $p_\phi$ on the network weights.

\subsection{Illustrative example}
First, we analyse our proposed hierarchical Bayesian model w.r.t. its ability to cope with the inherent ambiguity of different parametrizations leading to the same interventional but different counterfactual distributions by learning a BW-GP on data arising from the SCM of Eq.~\ref{eq:illustrative_example_scm} (see also Fig.~\ref{fig:illustration_ambiguous_counterfactual}). To also account for probing the learned model in not well covered regimes of the training data, we selected 174 training points all of which lying within $[0,0.6]$ but tested the model also in the regime $[0.6,1]$. On these training datapoints, we fitted both a BW-GP as well as a Gaussian Process. To assess the quality of the modelled SCM, we generated 1000 samples of $X_1$ uniformly across the range $[0,1]$ and draw one sample from the modelled interventional distribution. The resulting predictive distribution of the BW-GP and GP are illustrated in  
 \begin{wrapfigure}{r}{0.77\textwidth}
    \centering
    \subfigure[BW-GP]{
    \label{fig:wgp_interventional_dist_var_phi_var_eps}
    \includegraphics[width=0.46\linewidth,scale=0.5]{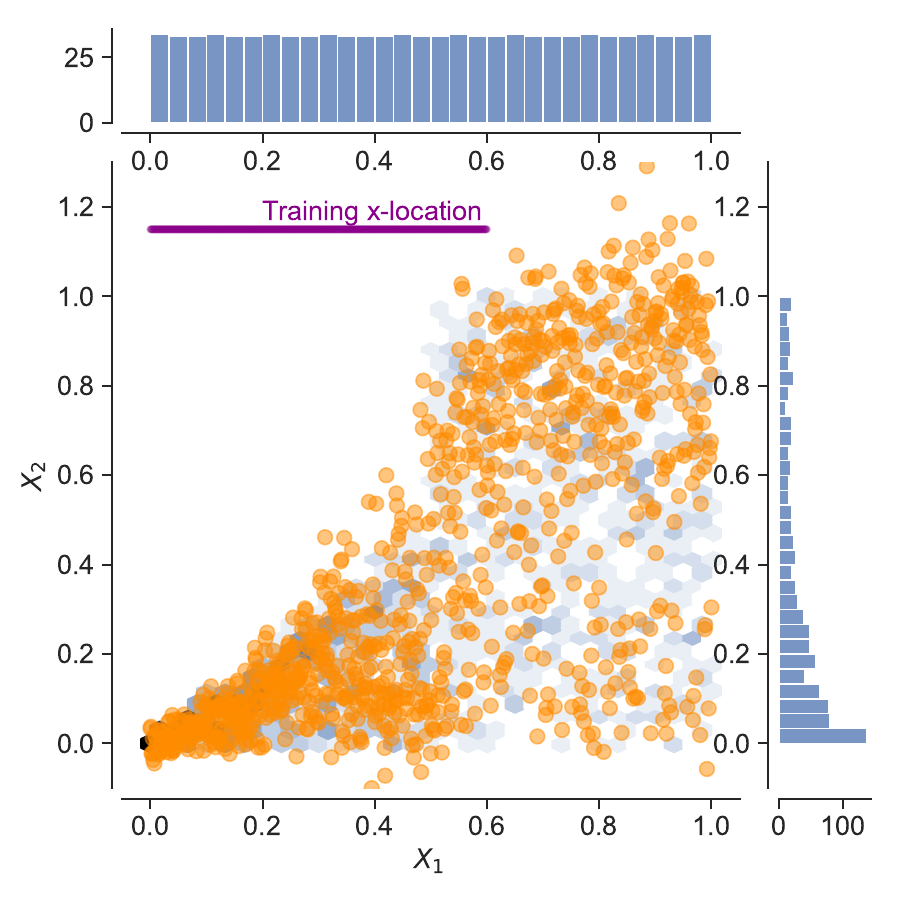}}\hfill 
    \subfigure[GP]{
    \label{fig:gp_interventional_dist_var_phi_var_eps}
    \includegraphics[width=0.46\linewidth,scale=0.5]{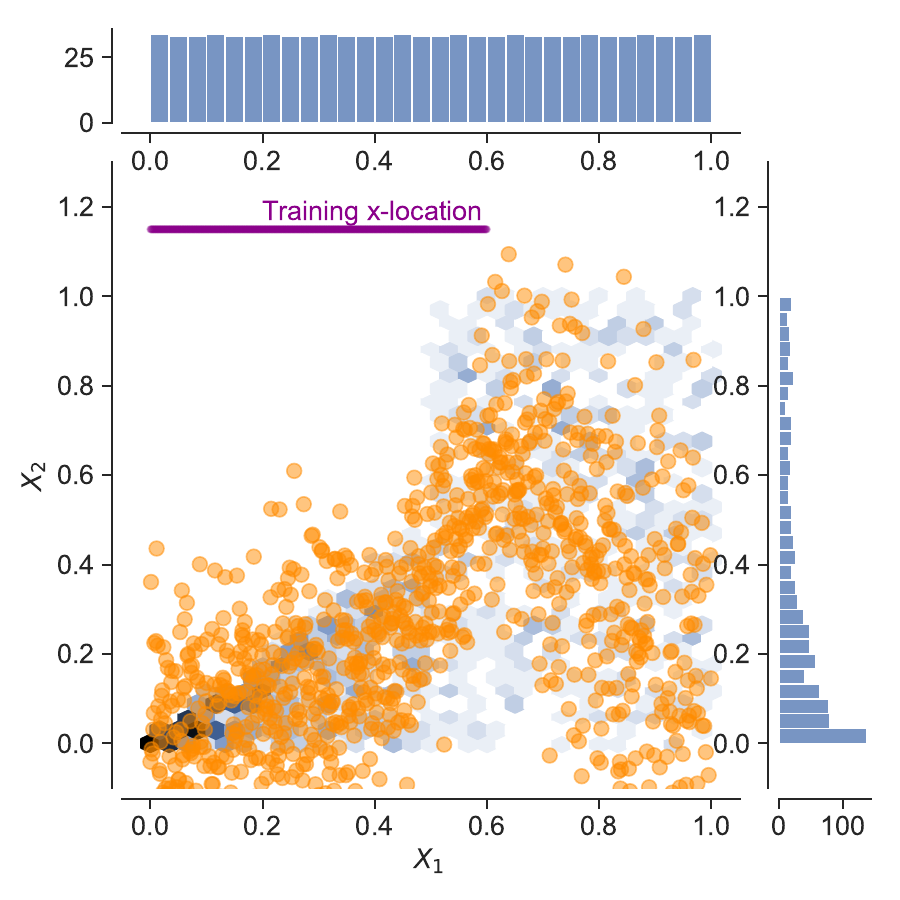}}
    \caption{Comparison of the modeled interventional distribution.}\vspace{-1.em}
    \label{fig:iv_gp_wgp_var_var}
\end{wrapfigure}
Fig.~\ref{fig:iv_gp_wgp_var_var}a and Fig.~\ref{fig:iv_gp_wgp_var_var}b respectively. Both models are trained on points between 0 and 0.6, rendering the range between 0.6 and 1 as extrapolation regime. The blue points in the background show the observational distribution of the SCM Fig~.\ref{eq:illustrative_example_scm}, the orange points correspond to  samples of the interventional distribution of (a) our BW-GP and (b) a GP. Blue points in the background indicate samples from the ground truth model of Eq.~\ref{eq:illustrative_example_scm} (see Fig.~\ref{fig:illustration_ambiguous_counterfactual}).
As can be seen from Fig.~\ref{fig:iv_gp_wgp_var_var}, the BW-GP provides a close fit to the ground truth observational distribution whereas a GP is not able to fit the observational data as accurately, due to the non-stationary noise distribution. This heteroscedasticity of the noise distribution also forces the plain GP  to explain the data using non-zero functional coupling uncertainty. The BW-GP model, however nicely adjusts for such uncertainty by allowing for non-stationary distributions over functional couplings.

\begin{wrapfigure}{r}{0.77\textwidth}
    \centering
    \subfigure[\small{BW-GP}]{
    \label{fig:cf_wgp_fix_fix}
    \includegraphics[width=0.47\linewidth]{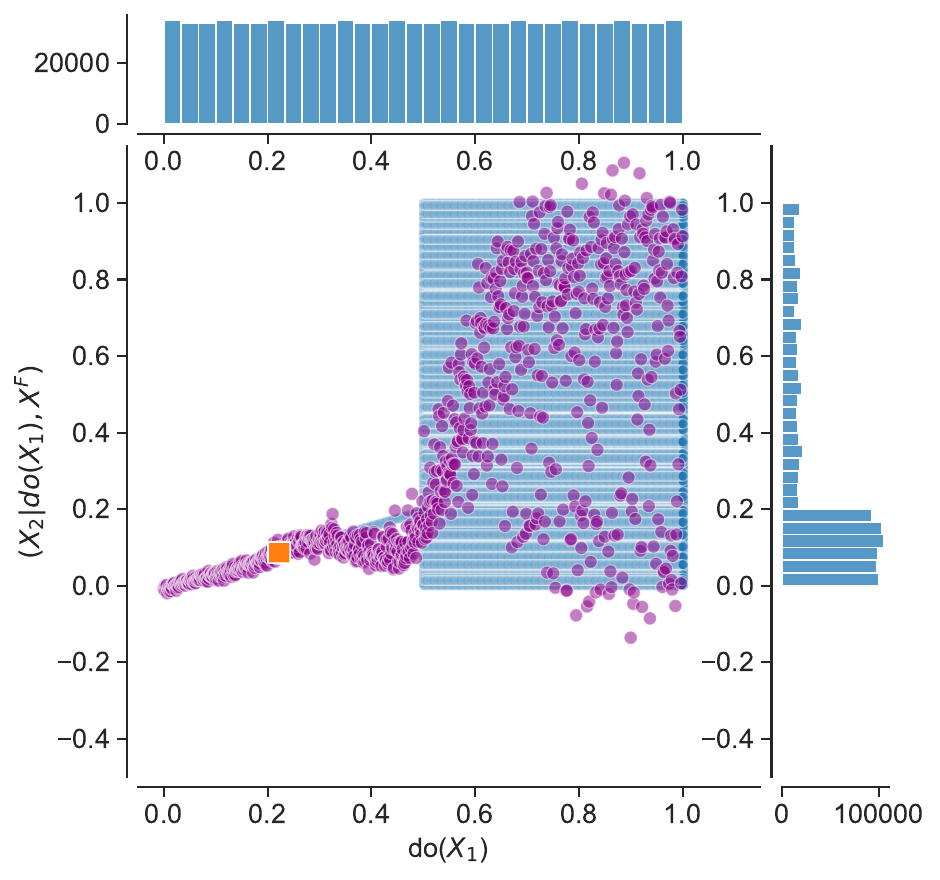}}\hfill
    \subfigure[\small{GP}]{
    \label{fig:counterfactual_gp}
    \includegraphics[width=0.47\linewidth]{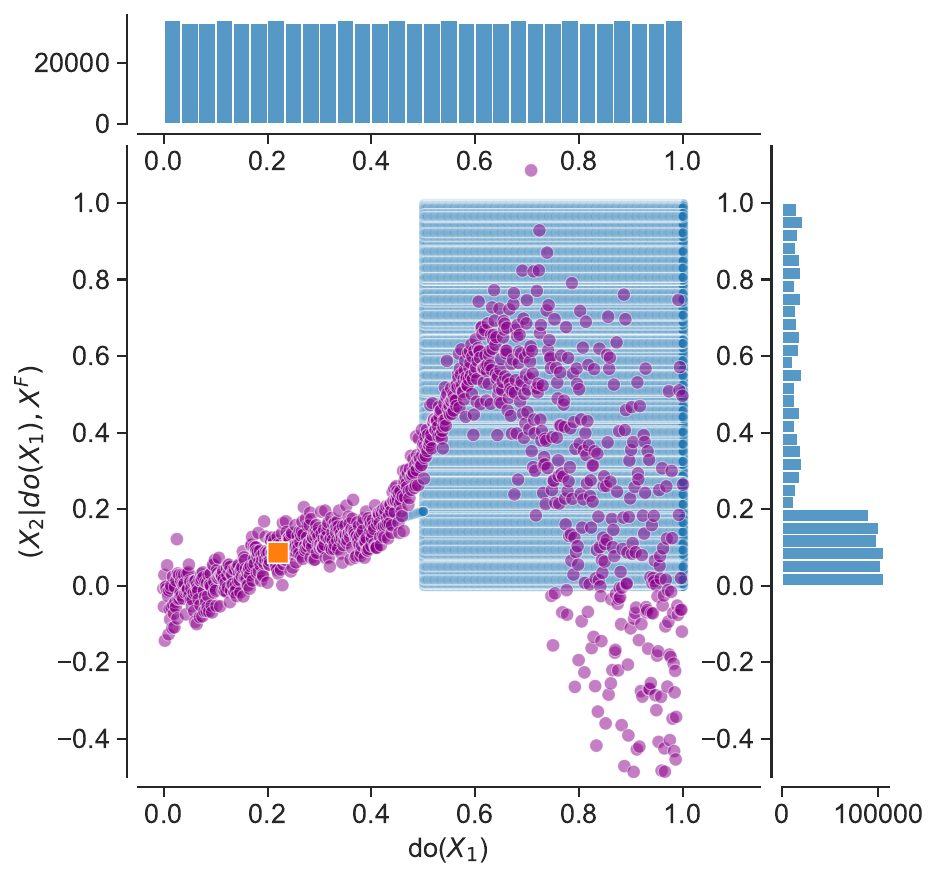}}
    \caption{Illustration of the modeled counterfactual distribution.} \vspace{-0.5em}
    \label{fig:counterfactual_comparison}
\end{wrapfigure}

Second, we also evaluate the counterfactual distribution for both a Gaussian Process without parametrization uncertainty and our BW-GP which includes such uncertainty. In Fig.~\ref{fig:counterfactual_comparison}, we plot the resulting counterfacutal distribution estimates when intervening on $X_1$ and using the noise-posterior of the observation $X^F_1=0.22, X^F_2=0.08$ (marked by an orange square in Fig.~\ref{fig:illustration_ambiguous_counterfactual}) and compare them against counterfactual distributions arising from different parametrization in Eq.~\ref{eq:illustrative_example_scm}. Here, the blue points in the background show samples of the true counterfactual distribution constructed from the factum (orange box) and varying parametrizations $\phi$. The purple points represent a sample drawn of the counterfactual distribution of (a) our BW-GP and (b) a GP.
The interventional distribution of the counterfactual SCM (as shown in Fig.~\ref{fig:counterfactual_comparison}) is forced to recover the observation that it is conditioned on, if we would intervene on $X_1$ forcing the variable to have the same value as observed (orange marker in Fig.~\ref{fig:counterfactual_comparison}). While this property is recovered by both BW-GP and GP (by construction of the counterfactual SCM), stationarity assumption of the noise of the GP results in larger uncertainty around the observation in the counterfactual. 
Despite the non-stationarity of the noise of the BW-GP, it seems to also cover the uncertainty of counterfactual distribution in the out-of-training data regime.
We focus on isolating the impact of uncertainties stemming from the inherent ambiguity of different parametrizations of the same observational and interventional distribution.

\subsection{Benchmark Experiments}\label{sec:benchmark_experiments}

Besides the illustrative example we evaluated the BW-GP on an important downstream task of a counterfactual distribution to assess the impact of the BW-GP on a more realistic decision making processes. 
To this end, we compare our model (BW-GP) against other baseline methods within  algorithmic recourse benchmark of \cite{Karimi-baseline}, including a standard GP, a linear regressor and a conditional variational autoencoder (CVAE). 
For the CVAE, we use the implementation of \cite{Karimi21}, yet it can be regarded as a non-amortized version of the CVAE by \cite{NEURIPS2020_0987b8b3}.
 Analogously to  \cite{Karimi-baseline}, we compared both the counterfactual model (denoted by  $\mathcal{M}_{\texttt{<model>}}$) as well as the interventional variants of the different models (denoted by $\text{CATE}_{\texttt{<model>}}$).
\begin{table*}[ht]
 \caption{Experimental results of a three variable causal model in a recourse setting with 100 individuals. We compare our model ($\mathcal{M}_\text{BW-GP}$) against the reproduced baselines $\texttt{LIN}, \texttt{GP}, \texttt{CVAE}$ of \cite{Karimi-baseline}.}
 \resizebox{1\textwidth}{!}{
\begin{tabular}{lccccccccl}\toprule
& \multicolumn{3}{c}{Linear SCM} & \multicolumn{3}{c}{NON LINEAR SCM} & \multicolumn{3}{c}{NON ADDITIVE SCM}
\\\cmidrule(lr){2-4}\cmidrule(lr){5-7}\cmidrule(lr){8-10}
           & $\texttt{Valid}(\%)$      & Cost($\%$)       & $\texttt{MMD}$
           & $\texttt{Valid}(\%)$      & Cost($\%$)       & $\texttt{MMD}$ 
           & $\texttt{Valid}(\%)$      & Cost($\%$)       & $\texttt{MMD}$ \\\midrule
$\mathcal{M}_{*}$           & 100 & 11.2 $\pm$ 7.4 & - & 100 & 19.7 $\pm$ 12.3 & - & 100 &  10.3 $\pm$ 8.6 & - \\
$\mathcal{M}_\text{LIN}$    & 100 & 12.0 $\pm$ 8.0 & 0.019 $\pm$ $2.37 \cdot 10^{-5}$ & 67   & 20.6 $\pm$ 10.8 & 0.202 $\pm$ 0.006   & 100 & 10.1 $\pm$ 8.3  & 0.383 $\pm$ 0.027 \\
$\mathcal{M}_\text{GP}$     & 100 & 13.3 $\pm$ 9.7 & 0.043 $\pm$ 0.001 & 100 & 22.0 $\pm$ 13.5 & 0.036 $\pm$ 0.001  & 98 &  10.3 $\pm$ 8.5 &  0.369 $\pm$ 0.019\\
$\mathcal{M}_\text{CVAE}$   & 100 & 12.7 $\pm$ 8.2 &  0.031 $\pm$ 0.001 &  91    & 25.4 $\pm$ 14.3  & 0.139 $\pm$ 0.002  & 97 &   10.1 $\pm$ 8.1  & 0.146 $\pm$ 0.013  \\
\rowcolor{lightgray} $\mathcal{M}_\text{BW-GP}$    & 100  & 13.0 $\pm$ 9.0 & 0.069 $\pm$ 0.002  & 99 & 22.3 $\pm$ 14.7  &  0.043 $\pm$ 0.001 & 99 & 10.2 $\pm$ 9.1  & 0.120 $\pm$ 0.009\\
$\text{CATE}_{*}$  & 88  & 12.4 $\pm$ 9.6 &  - & 99 & 28.1 $\pm$ 28.9 & - & 100 &  10.1 $\pm$ 8.2  & -\\
$\text{CATE}_\text{GP}$     & 90  & 12.6 $\pm$ 8.5 &  0.044 &  97  & 27.4 $\pm$ 17.8 & 0.043 & 94 & 9.6 $\pm$ 8.5 &   0.261\\
$\text{CATE}_\text{CVAE}$   & 87  & 12.8 $\pm$ 10.5 &  0.066 & 99 & 33.4 $\pm$ 25.0 & 0.069 & 100 &  10.1 $\pm$ 8.3 & 0.064\\
\rowcolor{lightgray} $\text{CATE}_\text{BW-GP}$ & 93  & 12.8 $\pm$ 9.0 & 0.073 & 98   & 29.8 $\pm$ 19.4  & 0.039  & 98 &   9.7 $\pm$ 7.9  & 0.089 \\
\bottomrule
\end{tabular}
\label{tb:results3var}}
\end{table*}

\begin{table*}[ht]
\begin{center}
 \caption{Experimental results of a seven variable semi synthetic causal model on 100 facta in a recourse setting.} 
 \resizebox{1\textwidth}{!}{
\begin{tabular}{lccccccccl}\toprule
& \multicolumn{3}{c}{LINEAR LOG. REGR.} & \multicolumn{3}{c}{NON-LINEAR LOG. REGR.} & \multicolumn{3}{c}{RANDOM FOREST}
\\\cmidrule(lr){2-4}\cmidrule(lr){5-7}\cmidrule(lr){8-10}
           & $\texttt{Valid}_{*}(\%)$      & Cost($\%$)      & $\texttt{MMD}$ 
           & $\texttt{Valid}_{*}(\%)$      & Cost($\%$)      & $\texttt{MMD}$ 
           & $\texttt{Valid}_{*}(\%)$      & Cost($\%$)      & $\texttt{MMD}$ \\\midrule
$\mathcal{M}_{*}$           & 100 & 17.4 $\pm$ 8.0  & - & 100   & 15.8 $\pm$ 9.3 & - & 100 &  19.3 $\pm$ 9.1 & -\\
$\mathcal{M}_\text{LIN}$    & 100  & 18.0 $\pm$ 8.3  & 0.121 $\pm$ 0.007 & 96 & 16.2 $\pm$ 9.5 & 0.101 $\pm$ 0.009 & 94  &  19.5 $\pm$ 9.4 & 0.094 $\pm$ 0.007\\
$\mathcal{M}_\text{GP}$     & 100 & 22.0 $\pm$ 8.7  & 0.128 $\pm$ 0.004 & 100   & 18.6 $\pm$ 10.4 & 0.042 $\pm$ 0.001 & 100 &  21.2 $\pm$ 9.4 & 0.040 $\pm$ 0.001\\
\rowcolor{lightgray} $\mathcal{M}_\text{BW-GP}$  & 100 & 22.3 $\pm$ 9.3  & 0.050 $\pm$ 0.002 & 100   & 19.6 $\pm$ 12.1  & 0.053 $\pm$ 0.002  & 99 & 20.7 $\pm$ 9.2  & 0.049 $\pm$ 0.002 \\
$\text{CATE}_{*}$           & 88  & 25.7 $\pm$ 9.3  & - & 89    & 21.4 $\pm$ 14.2 & - & 92  &  23.9 $\pm$ 9.0 & -\\
$\text{CATE}_\text{GP}$     & 91  & 26.6 $\pm$ 9.5  &  0.082 & 93    & 22.3 $\pm$ 14.8 & 0.088 & 98 &  24.5 $\pm$ 9.5 & 0.086\\
\rowcolor{lightgray} $\text{CATE}_\text{BW-GP}$ & 95  & 28.1 $\pm$ 11.1 & 0.090 & 94 & 22.5 $\pm$ 14.4  & 0.087  & 98 &   24.6 $\pm$ 9.4 & 0.077  \\
\bottomrule
\end{tabular}
\label{tb:results7var}}
\end{center}
\end{table*}

In this algorithmic recourse benchmark setting, the goal is to find both the optimal nodes for an intervention as well as the optimal intervention value in relation to the cost (Eq.~\ref{eq:algorithmic_recourse_uncertainty}). We report validity and cost of \cite{Karimi-baseline}, where the validity defines the percentage of individuals with a beneficial outcome after a counterfactual sample is drawn. The cost is the L2-norm between the factum $\factum$ and the intervention, normalised by the range of each training variable. 
To assess the quality with which we represent the counterfactual distribution, not just the algorithmic recourse task, we additionally, evaluate the maximum mean discrepancy (MMD) \citep{gretton2012kernel} between the modelled counterfactual distribution and the counterfactual distribution of the ground truth model (denoted as $\mathcal{M}_{*}$). As both depend on the observed factum, we average the obtained MMD values across 100 facta. 
More precisely, to generate a sample of the modeled counterfactual distribution, we first calculate the posterior noise distribution and then perform a soft intervention on the root node by sampling values for the root node from the ground truth distribution. Using this sampling process, we obtain one counterfactual sample per factum. The same sampling process is used to evaluate the quality of the modeled interventional distribution in terms of MMD value, however, the noise prior is used instead of the noise posterior per factum to generate a sample. To generate samples from counterfactual distribution of the ground truth SCM, we stored the noise variables $\mathbf{U}$ that generated a particular $\factum$ in the test data and substituting it in the structural equations of the SCM after performing an intervention. In order to use the same MMD metric across different models, we used a squared exponential kernel and used two independent samples of the ground truth distribution to estimate hyperparameters of the kernel according to the median heuristic \citep{garreau2017large}.

\paragraph{Synthetic three variable causal model}
First of all we evaluate our model on three SCMs, a linear and a non-linear both with additive noise and a nonlinear with non-additive noise. Each SCM has the same underlying causal graph consisting of three variables yet differs in the functional couplings being either linear, non-linear or exhibiting non-additive noise. Since the ground truth is known of this artificial, we can generate data from it. Analogously to \cite{Karimi-baseline}, we trained each model on 250 such samples from observational distribution and evaluated on 100 facta sampled from the observational distribution which are found to be negatively classified according to a logistic regression, see Tab.~\ref{tb:results3var}.  Here, $CATE_*$ refers to the optimization process in which interventions are evaluated w.r.t. the interventional SCM rather than the counterfactual SCM within Eq.~\ref{eq:algorithmic_recourse_uncertainty} (in the constraint set), see  \citep{Karimi-baseline}. Therefore, interventions found by $CATE_*$ in Tab.~\ref{tb:results3var},\ref{tb:results7var} are not necessarily achieving 100 percent validity when checked with the counterfactual ground truth SCM. To set hyperparameters of our model (number of bins in the spline and size of the neural network), we performed a Bayesian optimization on a validation set (details can be found in the suppl.~material). Although the BW-GP performs comparably in terms of costs and validity as the other best models, on the non-additive SCM we show a significantly smaller MMD than the GP in the counterfactual and interventional (CATE) task. This could be due to the fact that the normalizing flow is able to learn multimodal distributions well. Nevertheless the GP achieves high validity and comparable loss, which means that the learned conditional distributions do not have a strong impact on the recourse task itself. The conditional variational autoencover (CVAE) performs similarly well on the non additive SCM but operates considerably worse on the non linear SCM counterfactual task. As noted by \cite{Karimi-baseline} samples of $\mathcal{M}_{\texttt{CVAE}}$ are "pseudo-counterfactual" possible amounting to a reduced accuracy.

In the linear SCM experiment, we observe that the BW-GP performed slightly worse than the GP in terms of the MMD, yet without significant impact on validity or costs. 
Note, however, that the costs and validity are computed based on a counterfactual distribution which is constructed from a single ground truth SCM and hence does not include the additional uncertainty of potentially different parametrizations. 
We argue that the slight drop in performance of the MMD can therefore be attributed to the additional uncertainties accounted for by the BW-GP. Therefore, we additionally measured the variance of the counterfactual distribution samples over the different facta to assess a potential increase in the overall uncertainty of the counterfactual distribution modeled by the different methods. Indeed, we observed that our model has the highest variance (2.9907) across counterfactual distribution samples followed by the GP (2.9531), the linear model (2.9271) the CVAE (2.9189).



\paragraph{Semi synthetic seven variable causal model}
The semi synthetic seven variable system is inspired by the German Credit UCI dataset as it features relevant variables such as age, savings, gender etc. as well as a labelling mechanism representing the loan-approval. Based on data generated from this constructed SCM, different classifiers are trained: linear and non-linear logistic regression, and a random forest. Similarly to the three-variable model we used the same benchmark setting, models and computation as in \cite{Karimi-baseline} and performed a hyper-parameter optimization on a validation set. Also in this more realistic and higher dimensional setting, we observe a more accurate characterization of the counterfactual distribution as indicated by significantly lower MMD scores without sacrificing validity (see Tab.~\ref{tb:results7var})\footnote{Note that the results for the variational autoencoder could not be reproduced with the provided source code.}. While the BW-GP performs slightly worse than the GP in terms of accuracy of the interventional distribution for the logistic regression setting, it still achieves better validity. 
Similarly to the evaluation within the three-variable model, each method is only evaluated against a single SCM assumed to be the ground truth. However, the BW-GP additionally accounts for the uncertainty in the parametrization leading to larger spread of counterfactual costs as indicated by the standard errors, yet without sacrificing validity.

\section{Conclusion}\label{sec:conclusion}
In this paper, we proposed a hierarchical Bayesian model to account for ambiguities in the underlying SCM as well as for the uncertainties arising from imperfect knowledge of functional couplings due to limited observational data. By using a Bayesian Warped GP, we were able to not only allow for non-Gaussian distribution at descendent nodes, but also non-stationary noise distributions. 
This seems to be particularly beneficial for counterfactual distributions (see Figure~\ref{fig:counterfactual_comparison}). Although we introduced an additional source of uncertainty about the parametrization, this resulted in a more accurate fit of the counterfactual distribution also in more realistic settings (see Table.~\ref{tb:results3var}\ref{tb:results7var}).

The gained expressiveness of the model also leads to robust recourse actions in terms of the achieved validity without an increase in costs due to the additional uncertainty within considered SCMs. In fact, our BW-GP \ref{eq:generative_model} theoretically provides a sufficiently flexible model to capture any conditional distribution $p(X_r|X_{\rm{pa}(r)})$. However, in practice the flexibility of the neural network as well as the amount of observational data is limited. In this limited case, the ground truth models of sec.(4) will not be exactly matched by our model. Therefore, our experiments can be considered as evaluation results under model misspecification.
The proposed method can also be used in settings with unobserved confounders by introducing additional, yet unobserved nodes within the SCM and integrating out their values during the training phase. However, when falsely assuming potential hidden confounders by introducing latent variables, each of which are associated with a flexible probability distribution, predictive power is likely to decline.
Although we have shown that the proposed model can account for these ambiguities to a certain degree, it still contains hard and soft assumptions one can relax. For example, in this research we assumed that the graphical structure between the static modeled variables is known. By imposing yet another probability distribution on the graphical structure, such a hard assumption can be relaxed with the downside of additional computational complexity to learn these models \citep{von2019optimal}.

\bibliography{acml23}    



\newpage

\section{SUPPLEMENTARY MATERIAL}
\subsection{HYPERPARAMETER SETUP}\label{apd:hp-setup}
Since our Bayesian Warped GP consists of various components, there are a couple of hyperparameters that can be optimized, see Table~(\ref{tb:hyperpara}). 
We represent the bijective mapping $g_{\phi}$ of the Normalizing flow by a neural spline flow transform with element-wise (referred to as bins) rational conditional spline functions, to represent the conditional distributions in the causal model.
Each conditional spline transform consists of a dense neural network with three Bayesian linear layers and a RELU activation function. 
In this neural network the hidden dimensions ($\texttt{hidden dims}$) need to be set (implemented with Pyro \citep{bingham2019pyro}). 
Furthermore, the spline is defined in a bounding box ($\texttt{bounds}$), which should cover the range of input data, for details see \cite{durkan2019neural}. To relax this requirement, we normalize the input data. \\
According to our variational inference scheme, we can optimize further parameters affecting the training: the number of monte carlo samples ($\texttt{S}$) to be drawn, the prior variance ($\texttt{prior var}$), the learning rate ($\texttt{lr}$), and the training steps ($\texttt{steps}$). We optimize these variables in the seven variable setup to minimize the MMD on a held-out validation dataset of size 250 (generated from the ground truth SCM). For the three variable setup we optimized the hyperparameters w.r.t. the cost due to time-constraints. In both cases, we used the BOHB (Bayesian Optimization algorithm using Hyperband) algorithm \citep{DBLP:journals/corr/abs-1807-01774} for optimization. More specifically, we used the python ray-tune package of \cite{liaw2018tune} as implementation of BOHB.
\begin{table*}[ht]
\begin{center}
 \caption{Optimal Hyperparameters found with BOHB on a validation set for each SCM and classifier setting.} 
 \resizebox{1\textwidth}{!}{
\begin{tabular}{lccccccccl}\toprule
& \multicolumn{1}{c}{LINEAR SCM} & \multicolumn{1}{c}{NON-LINEAR SCM} & \multicolumn{1}{c}{NON-ADDITIVE} & \multicolumn{1}{c}{LINEAR LOG. REGR.} & \multicolumn{1}{c}{NON-LINEAR LOG. REGR.} & \multicolumn{1}{c}{RANDOM FOREST}\\
\\\cmidrule(lr){2-2}\cmidrule(lr){3-3}\cmidrule(lr){4-4}\cmidrule(lr){5-5}\cmidrule(lr){6-6}\cmidrule(lr){7-7}
$\texttt{bounds}$           & 6 & 1 & 10 & 27 & 3 & 21\\
$\texttt{hidden dims}$    & 10 & 13 & 40 & 2 & 6 & 27\\
$\texttt{lr}$     & 0.03 & 0.03 & 0.01 & 0.04 & 0.008 & 0.05\\
$\texttt{steps}$  & 5719 & 5719 & 4501 & 6982 & 6198 & 4956\\
$\texttt{S}$          & 15 & 21 & 20 & 31 & 24 & 21\\
$\texttt{prior var}$    & 0.1 & 0.1 & 0.05 & 0.03 & 0.01 & 0.02\\
\bottomrule
\end{tabular}
\label{tb:hyperpara}}
\end{center}
\end{table*}
\subsection{INTERVENTIONAL DISTRIBUTION}
\begin{figure*}[ht!]
    \centering
    \subfigure{
    \label{fig:model-fit-ground-truth}    
    \includegraphics[width=0.45\linewidth]{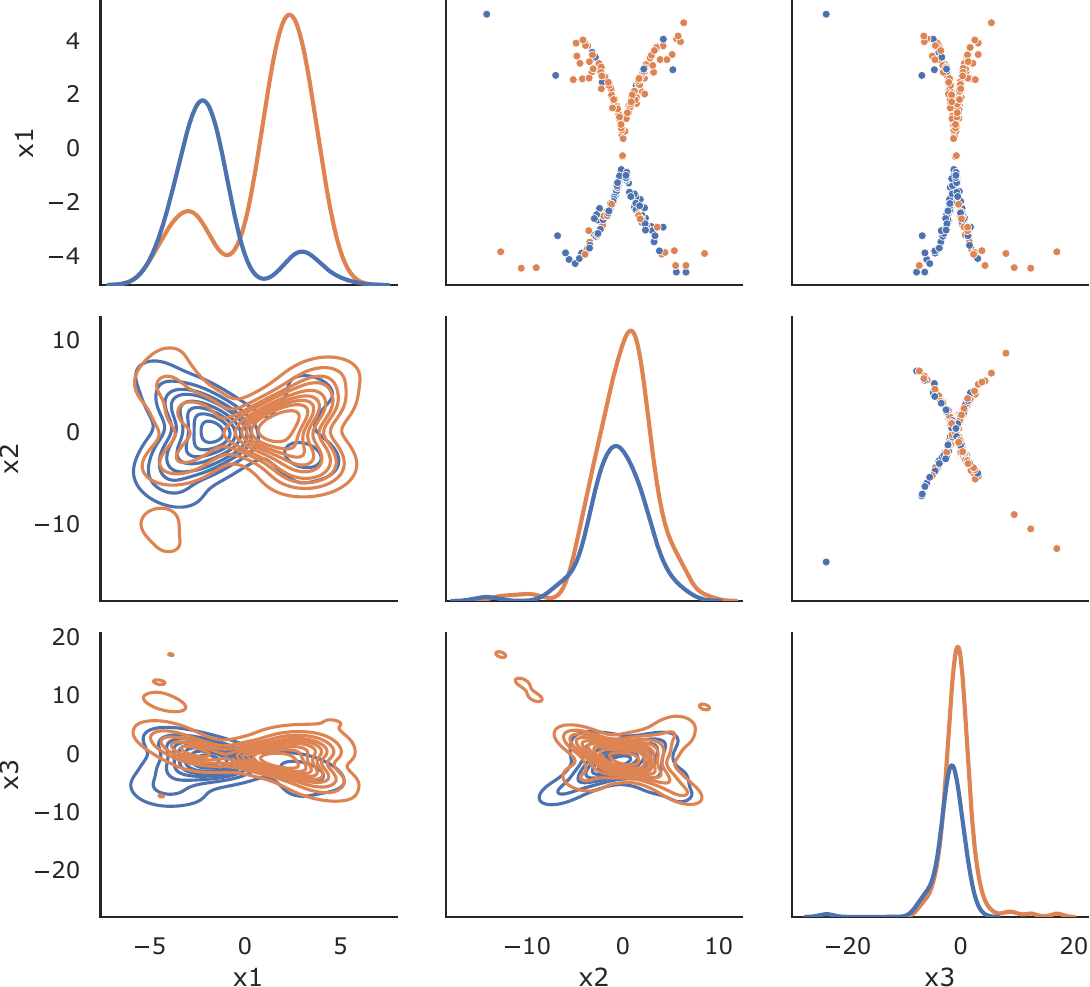}}\hfill 
    \subfigure{
    \label{fig:our-model}\hfill
    \includegraphics[width=0.45\linewidth]{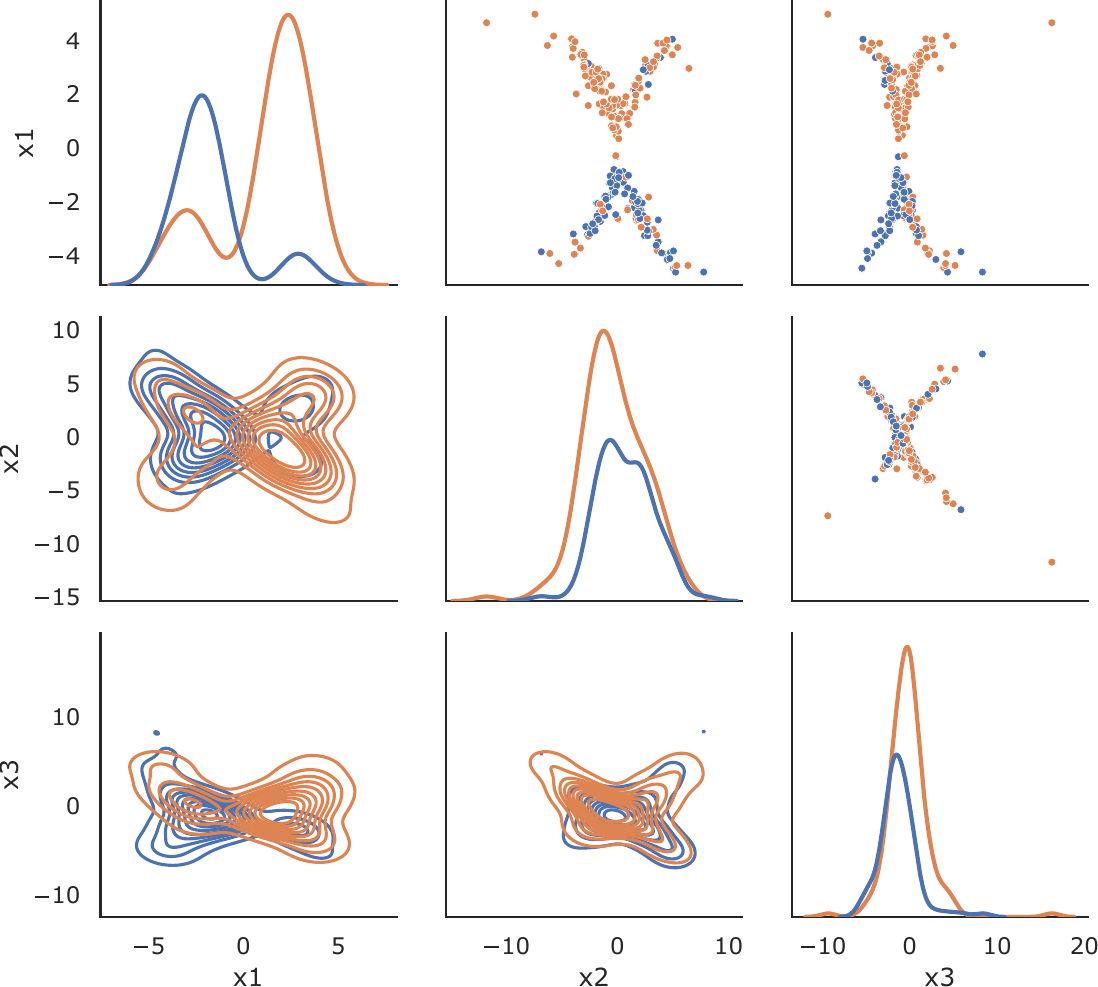}}
    \caption{In (a) we plot the ground truth and in (b) the Bayesian Warped Gaussian Process model. The coloring corresponds to the classes the classifier yields in the recourse task (blue are the negatively and orange the positive classified points).}
    \label{fig:model_fit-ground-bwgp}
\end{figure*}

\begin{figure*}[ht!]
    \centering
    \subfigure{
    \label{fig:model-fit-gp}    
    \includegraphics[width=0.45\linewidth]{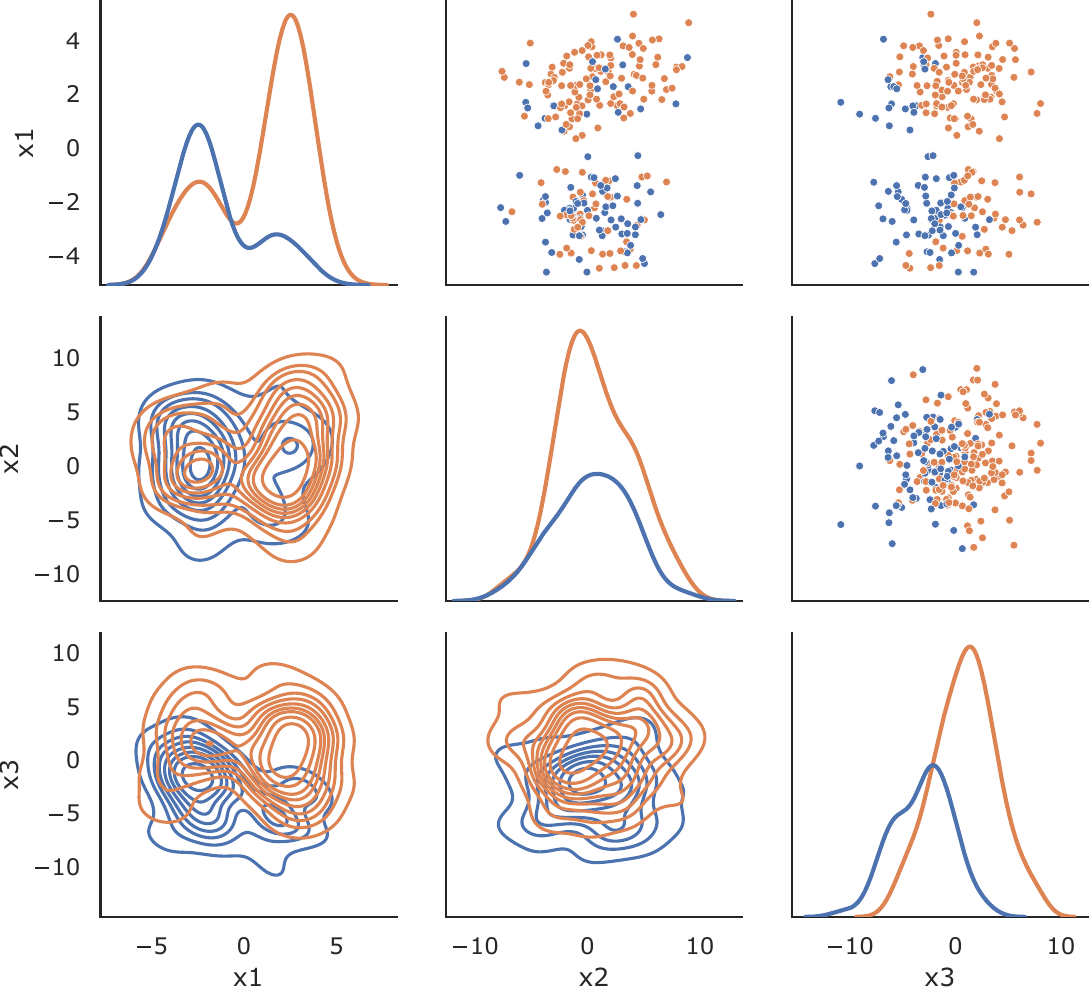}}\hfill
    \subfigure{
    \label{fig:model-fit-cvae}
    \includegraphics[width=0.45\linewidth]{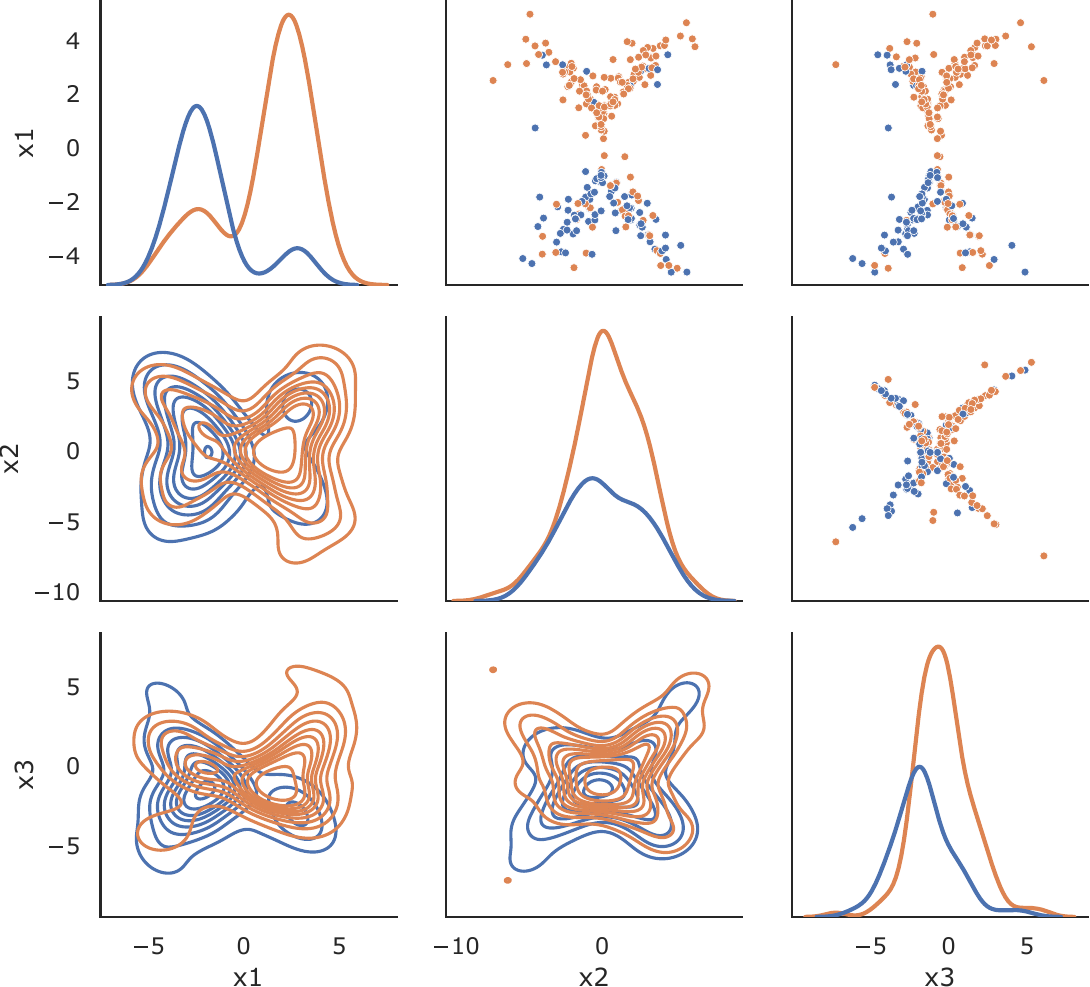}}
    \caption{In (a) the Gaussian Process is plotted and in (b) the CVAE model. The coloring corresponds to the classes the classifier yields in the recourse task (blue are the negatively and orange the positive classified points).}
    \label{fig:model_fit-cvae-gp}
\end{figure*}

In this section, we provide additional plots, illustrating the properties of the different models visually. Due to the complex yet low-dimensional setting, the non-additive SCM of the three-variable model (see Tab.1) is of particular interest. Since this was visually not the case for the other SCMs, we do not explicitly show them. In Fig.~\ref{fig:model_fit-ground-bwgp} we plotted the ground truth distribution (see Fig.~\ref{fig:model_fit-ground-bwgp}a) and in Fig.~\ref{fig:model_fit-ground-bwgp}b the corresponding distribution as modeled by the BW-GP. To generate samples from the different models, we generated samples from the ground truth model for the variable of the root-node $X_1$. Using the different models for the conditional distributions of children given the parents, we generated the remaining variables $X_2,X_3$ according to the causal graph.
As also indicated by the small MMD-Values (cf. Tab.1), the BW-GP also visually fits the ground truth data much better than the GP (see Fig.~\ref{fig:model_fit-cvae-gp}a), which learns two Gaussian distributions for the multimodal distribution. While the CVAE in Fig.~\ref{fig:model_fit-cvae-gp}b fits the data also better than the GP, it exhibits a higher variance than the BW-GP, which is also reflected by a slightly larger MMD-value.

\end{document}